%% file: main.tex
\documentclass[10pt,twocolumn,letterpaper]{article}
 \usepackage{cvpr}              

\PassOptionsToPackage{numbers}{natbib}

\input{preamble}
\usepackage{booktabs}
\usepackage{multirow}
\usepackage{graphicx}
\usepackage{amsmath}
\usepackage{amssymb}
\usepackage{xcolor}
\usepackage{colortbl}
\definecolor{cvprblue}{rgb}{0.21,0.49,0.74}
\usepackage[pagebackref,breaklinks,colorlinks,linkcolor=blue,citecolor=blue,urlcolor=blue]{hyperref}

\usepackage[accsupp]{axessibility} 

\def\paperID{MTF-13} 
\def\confName{CVPR}
\def\confYear{2026}
\title{\textbf{MyoVision: A Mobile Research Tool and NEATBoost-Attention Ensemble Framework for Real Time Chicken Breast Myopathy Detection}
}
\author{Chaitanya Pallerla$^{1,2}$ \quad Siavash Mahmoudi$^{1}$ \quad Dongyi Wang$^{1,2,*}$\\[6pt]
$^{1}$~Department of Biological and Agricultural Engineering, University of Arkansas, Fayetteville, AR\\
$^{2}$~Department of Food Science, University of Arkansas, Fayetteville, AR\\
{\tt\small dongyiw@uark.edu}
}
\begin{document}
\maketitle
\input{sec/0_abstract}
\input{sec/1_intro}
\input{sec/2_methods}

\input{sec/3_results}
\input{sec/4_discussion}
\input{sec/5_conclusion}
{
    \small

\input{main.bbl}
}
\end{document}

%% file: preamble.tex
\PassOptionsToPackage{numbers}{natbib}
\usepackage[accsupp]{axessibility} 

%% file: sec/0_abstract.tex
\begin{abstract}
Woody Breast (WB) and Spaghetti Meat (SM) myopathies significantly impact poultry meat quality, yet current detection methods rely either on subjective manual evaluation or costly laboratory-grade imaging systems. We address the problem of low-cost, non-destructive multi-class myopathy classification using consumer smartphones. MyoVision is introduced as a mobile transillumination imaging framework in which 14-bit RAW images are captured and structural texture descriptors indicative of internal tissue abnormalities are extracted. To classify three categories (Normal, Woody Breast, Spaghetti Meat), we propose a NEATBoost-Attention Ensemble model, which is a neuroevolution-optimized weighted fusion of LightGBM and attention-based MLP models. Hyperparameters are automatically discovered using NeuroEvolution of Augmenting Topologies (NEAT), eliminating manual tuning and enabling architecture diversity for small tabular datasets. On a dataset of 336 fillets collected from a commercial processing facility, our method achieves 82.4\% test accuracy (F1 = 0.83), outperforming conventional machine learning and deep learning baselines and matching performance reported by hyperspectral imaging systems costing orders of magnitude more.
Beyond classification performance, MyoVision establishes a reproducible mobile RGB-D acquisition pipeline for multimodal meat quality research, demonstrating that consumer-grade imaging can support scalable internal tissue assessment.

\end{abstract}

%% file: sec/1_intro.tex
\section{Introduction}
\label{sec:intro}
Woody Breast (WB) and Spaghetti Meat (SM) are structural myopathies affecting poultry muscle tissue, leading to altered texture, reduced water-holding capacity, and significant economic loss in commercial processing~\cite{kuttappan2016white, soglia2016histology, tijare2016meat}. These defects primarily manifest as internal structural abnormalities rather than visible surface features, making automated detection challenging. In industrial processing lines, both conditions are identified through manual palpation or visual inspection after fillet separation approaches that suffer from limited reproducibility, operator fatigue, and inconsistent grading across facilities. Existing automated approaches depend on laboratory-grade modalities such as hyperspectral imaging~\cite{pallerla2024naswd}, near-infrared spectroscopy~\cite{wold2019near}, or custom machine vision systems~\cite{yoon2022development}. While effective, these systems require specialized hardware and controlled acquisition environments that fundamentally limit scalability and deployment flexibility in real processing settings~\cite{usda_nass_2023}. None provide a unified framework integrating deployable sensing, automated model optimization, and multimodal data acquisition within a single practical platform.

Transillumination imaging analyzes light transmitted through tissue to reveal subsurface structural variations that are not visible in conventional surface imaging, and has demonstrated utility in detecting internal defects in fruits and other agricultural products~\cite{han2024non, blasco2003machine, huang2015development}. In this study, the same principle is applied to the classification of poultry breast conditions into three categories: Normal, Woody Breast, and Spaghetti Meat, using a smartphone-based transillumination imaging system. Rather than reconstructing volumetric optical properties as performed in advanced biomedical imaging techniques such as diffuse optical tomography (DOT) or optical coherence tomography (OCT)~\cite{zacharopoulos2009dot, bowker2024oct, ekramirad2024oct}, the proposed approach captures a 2D spatially integrated attenuation pattern produced as broadband white light propagates through the chicken fillet. This attenuation pattern encodes aggregate variations in tissue density, fibrotic hardening, and fluid redistribution within the muscle structure. Woody breast and spaghetti meat myopathies exhibit fundamentally different structural characteristics, where woody breast is associated with fibrotic hardening and fluid accumulation, while spaghetti meat is characterized by connective tissue fragmentation and muscle fiber dissociation~\cite{soglia2016histology, chatterjee2016instrumental}. These structural differences generate distinct macroscopic light attenuation patterns that can be captured through transillumination imaging. To leverage these patterns, a computational pipeline is developed that integrates RAW image acquisition, gradient-based statistical features, frequency-domain texture descriptors, and ensemble machine learning models optimized through neuroevolution. This framework enables effective classification of poultry breast conditions using a low-cost imaging setup while retaining meaningful structural information relevant to muscle abnormalities.

Despite recent advances in machine learning for food quality inspection, existing studies rely on conventional classifiers with predefined architectures requiring extensive manual tuning~\cite{geronimo2019computer, munozlapeira2025hyperspectral, bergstra2012random, snoek2012practical}, which may fail to capture complex structural variations in myopathic tissue while remaining susceptible to local optima on discrete, heterogeneous feature distributions~\cite{goodfellow2016deep, pascanu2013difficulty}. Neuroevolution of Augmenting Topologies (NEAT) addresses these limitations by simultaneously evolving network topology and weights from minimal structures, augmenting complexity only when beneficial~\cite{stanley2002evolving, stanley2019designing}, eliminating manual architecture design and gradient-based assumptions. Although neuroevolution has shown strong potential in architecture search and reinforcement learning~\cite{real2019regularized, stanley2019designing}, its application in food quality evaluation remains largely unexplored. Therefore, this study explores a neuroevolution-based ensemble learning framework to improve multi-class poultry myopathy classification using smartphone-based transillumination imaging.

We propose \textbf{MyoVision}, a smartphone-based transillumination imaging and automated ensemble learning framework for myopathy classification from small, heterogeneous tabular feature spaces. A domain-specific pipeline extracts 16 structural descriptors quantifying gradient statistics, frequency-domain texture responses, and dense tissue characteristics. For classification, we introduce the \textbf{NEATBoost-Attention Ensemble}, combining LightGBM with attention-based multilayer perceptrons under NEAT-driven optimization~\cite{stanley2002evolving, stanley2019designing}, with optimized probability fusion leveraging complementary decision boundaries across heterogeneous base learners. Beyond classification, MyoVision integrates 14-bit RAW capture, LiDAR-based 3D point cloud acquisition, mesh reconstruction~\cite{kazhdan2013screened}, SAM-based segmentation~\cite{kirillov2023segment}, and ChatGPT-assisted analysis~\cite{openai2023gpt4} as a unified mobile research platform for standardized multimodal RGB-D dataset collection.

The main contributions of this work are:
\begin{itemize}
    \item A smartphone transillumination imaging framework capturing 2D spatially-integrated attenuation patterns for internal myopathy assessment without specialized optical hardware.
    \item A neuroevolution-optimized ensemble (NEATBoost-Attention) evolving network topology and hyperparameters via NEAT for automated model discovery on small, heterogeneous datasets.
    \item A unified mobile research platform integrating RAW imaging, LiDAR 3D acquisition, mesh reconstruction, Auto Segmentation and AI-assisted analysis for standardized multimodal RGB-D data collection and chicken breast myopathy classification.
\end{itemize}

The rest of the paper is organized as follows. \Cref{sec:related} reviews related work. \Cref{sec:methods} describes the MyoVision architecture, feature extraction pipeline, and classification methodology. \Cref{sec:results} presents classification results. \Cref{sec:discussion} contextualizes findings and discusses limitations, and \Cref{sec:conclusion} concludes the paper.

\section{Related Work}
\label{sec:related}

Recent advances in mobile computer vision have enabled smartphone-based diagnostic tools across agricultural domains. GranoScan~\cite{dainelli2024granoscan} deployed smartphone imagery for detection of over 80 wheat pests and diseases, achieving 77--95\% accuracy under real-world conditions. AgroAId~\cite{reda2022agroaid} applied EfficientNet via TensorFlow Lite for multi-class plant disease classification with regional outbreak analytics. Iftikhar et al.~\cite{iftikhar2024plant} and Trivedi et al.~\cite{trivedi2025cropleafnet} proposed CNN-based mobile apps for early disease detection, while Avci et al.~\cite{avci2025lightweight}, Niaz et al.~\cite{niaz2025efficient}, and Goklani~\cite{goklani2024realtime} demonstrated lightweight edge-deployable frameworks for real-time plant disease inference. While these studies confirm the viability of smartphone-based agricultural diagnostics, all are single-purpose classification tools targeting surface-visible symptoms without provision for research-grade data acquisition or multimodal sensor integration.

Smartphone-based meat quality assessment remains comparatively underdeveloped, with existing studies focused on narrow, task-specific surface attribute evaluation. Menezes et al.~\cite{menezes2025empowering} classified beef and pork tenderness from smartphone RGB images using deep neural networks, achieving F1-scores of 76.6\% and 81.5\% respectively. Lin et al.~\cite{lin2025smartphone} distinguished artificially marbled beef from premium Wagyu using GLCM and HSV features, achieving 94\% binary accuracy with Random Forest. In poultry, smartphone applications have focused exclusively on live-bird disease diagnosis: Degu et al.~\cite{degu2023smartphone} applied fecal imaging with YOLO~v3 and ResNet50 for Coccidiosis, Salmonella, and Newcastle Disease classification. A comprehensive review by Kalita et al. ~\cite{kalita2025application} covering 127 AI-based poultry monitoring studies identified no smartphone application addressing post-mortem structural meat quality, while Okinda et al. ~\cite{okinda2020review} highlighted variable illumination and annotation cost as persistent deployment barriers. Critically, none of these systems across plant or meat domains function as research platforms supporting multimodal sensor fusion or standardized data acquisition for longitudinal studies.

Automated detection of chicken breast myopathies has primarily relied on industrial imaging or spectroscopic hardware. Geronimo et al.~\cite{geronimo2019computer} combined computer vision with near-infrared spectroscopy achieving 91.8\% binary woody breast classification accuracy, while Yoon et al.~\cite{yoon2022development} developed an online bending-angle imaging system reporting 97.4\% two-class accuracy. Mu{\~n}oz-Lapeira et al.~\cite{munozlapeira2025hyperspectral} employed VIS--NIR hyperspectral imaging with SVM for three-class myopathy classification at 76.1\% accuracy, and texture- and morphology-based RGB imaging approaches have been explored for white striping and woody breast grading~\cite{caldascueva2021image, kato2019white, olaniyi2023structured}. Pallerla et al.~\cite{pallerla2024naswd} proposed NAS-WD, combining 2D transillumination with 3D LiDAR point cloud features for woody breast severity grading at 95\% accuracy, establishing that multimodal fusion improves fine-grained discrimination. Despite strong performance, none of these systems address three-class type classification (Normal, Woody Breast, Spaghetti Meat) using consumer-grade mobile sensing, nor provide research-grade multimodal data acquisition capability.

%% file: sec/2_methods.tex
\section{Methodology}
\label{sec:methods}
\begin{figure*}[t]
  \centering
  \includegraphics[width=\linewidth]{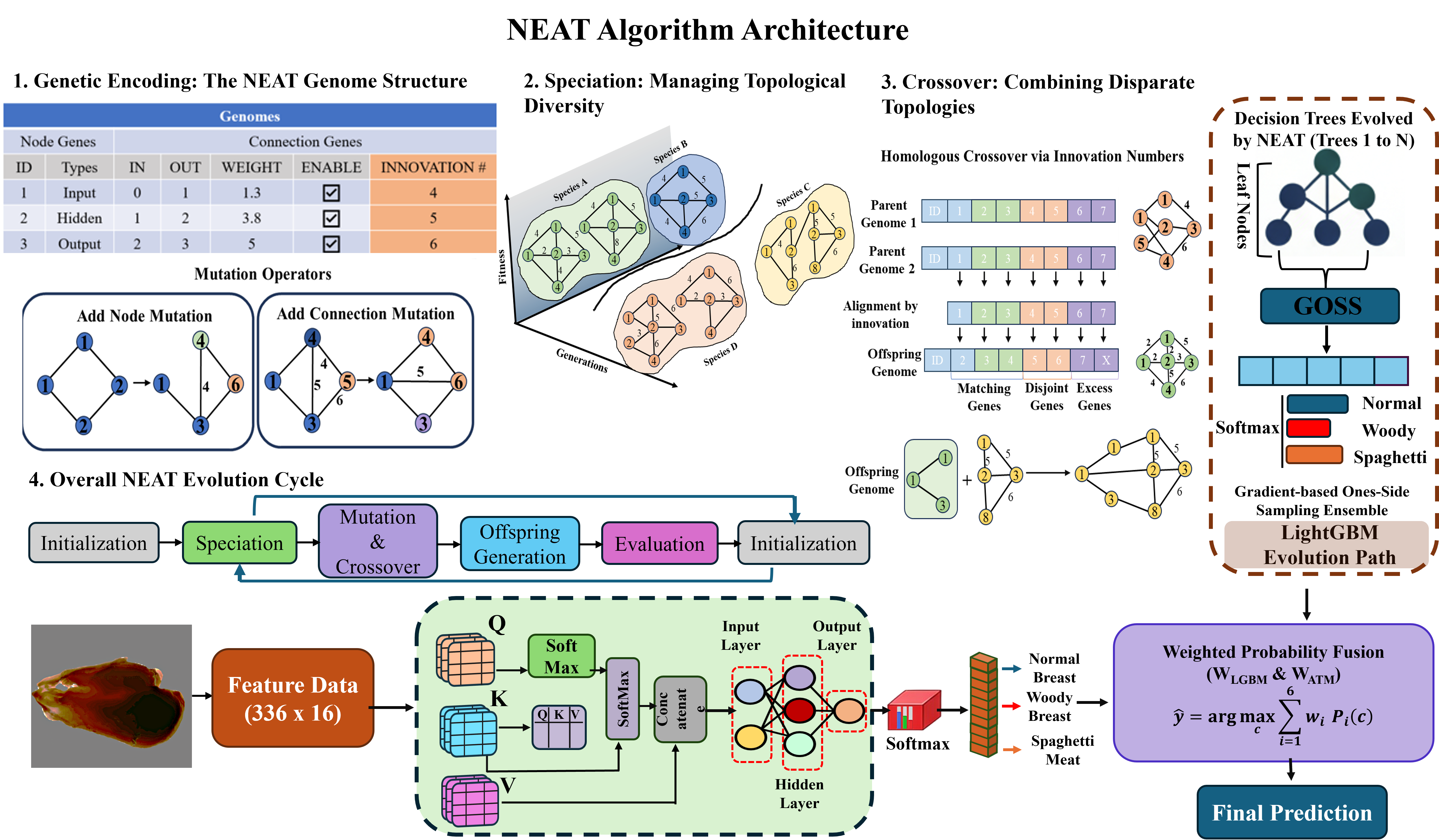}
  \caption{
Overview of the proposed NEATBoost-Attention ensemble framework. 
Transillumination images are converted into handcrafted structural descriptors, which are used to train candidate models. 
A NEAT evolutionary search generates optimized hyperparameters for both LightGBM and AttentionMLP classifiers. 
Predictions from the evolved models are aggregated through weighted probability fusion to produce the final myopathy classification.
}
  \label{fig:neat_arch}
\end{figure*}

This work introduces a neuroevolution-driven ensemble learning framework for smartphone-based poultry myopathy classification. The pipeline consists of three stages: (1) transillumination image acquisition, (2) extraction of handcrafted structural descriptors, and (3) neuroevolution-guided model optimization. Two complementary classifiers, a gradient boosting decision tree model (LightGBM) and an attention-based multilayer perceptron (AttentionMLP), are optimized using NeuroEvolution of Augmenting Topologies (NEAT)~\cite{stanley2002evolving, neatpython, stanley2019designing}. The resulting models are combined through weighted probability fusion.

As illustrated in Fig.\ref{fig:neat_arch}, the framework begins with the extraction of structural descriptors from transilluminated fillet images, producing a compact tabular feature representation. These features are used to train multiple candidate models whose hyperparameters are generated through the NEAT evolutionary search process. The NEAT module iteratively evolves populations of genome-encoded networks through speciation, mutation, and crossover to discover high-performing model configurations. The evolved hyperparameters are then used to instantiate both LightGBM and AttentionMLP classifiers. Finally, predictions from the optimized models are aggregated through weighted probability fusion to produce the final myopathy classification.

\subsection{NEATBoost-Attention Model Architecture}
\label{sec:classification}

The proposed classifier integrates two complementary learners: a gradient-boosted decision tree model (LightGBM) and an attention-based multilayer perceptron (AttentionMLP). Hyperparameters for both models are automatically discovered using NEAT-based neuroevolution. The resulting models are combined via weighted probability fusion to produce the final prediction.

Hyperparameter optimization is formulated as an evolutionary search problem. Each individual in the NEAT population encodes a small neural network that generates candidate hyperparameter configurations. Given a fixed input vector $z$, the genome network produces a normalized hyperparameter vector $h = f_{\theta}(z)$, where $h \in [0,1]^d$ denotes the normalized outputs. Each hyperparameter value is mapped to a valid range using

\begin{equation}
\lambda_i = l_i + h_i (u_i - l_i)
\end{equation}

where $[l_i,u_i]$ denotes the allowable range for hyperparameter $i$.

The fitness of each genome is evaluated using stratified cross-validation:

\begin{equation}
F(\theta) = \frac{1}{K} \sum_{k=1}^{K} \text{F1}_w^{(k)}
\end{equation}

where $\text{F1}_w^{(k)}$ denotes the weighted F1-score obtained on fold $k$. Evolution proceeds through mutation, crossover, and speciation operations over multiple generations, allowing networks to gradually increase in complexity while maintaining population diversity. The overall training procedure of the proposed NEATBoost-Attention framework is summarized in Algorithm~\ref{alg:neatboost}.
\begin{algorithm}[t]
\caption{NEATBoost-Attention Training Procedure}
\label{alg:neatboost}
\begin{algorithmic}[1]
\STATE Initialize NEAT population with random genomes
\FOR{generation $g = 1$ to $G$}
    \FOR{each genome $\theta$ in population}
        \STATE Generate candidate hyperparameters $h = f_{\theta}(z)$
        \STATE Map normalized outputs to valid ranges $\lambda_i = l_i + h_i(u_i-l_i)$
        \STATE Train candidate models (LightGBM, AttentionMLP) using $\lambda$
        \STATE Evaluate fitness using $K$-fold cross-validation
        \STATE Compute fitness score $F(\theta)$ as weighted F1
    \ENDFOR
    \STATE Apply NEAT evolutionary operations:
    \STATE \quad Selection based on fitness
    \STATE \quad Crossover between genomes
    \STATE \quad Mutation and speciation
\ENDFOR
\STATE Select top-performing models from evolved population
\STATE Optimize ensemble weights using Nelder–Mead simplex
\STATE Compute final ensemble prediction using weighted probability fusion
\end{algorithmic}
\end{algorithm}

The following sections describe the two base learners optimized through the NEAT search procedure.

The first learner is LightGBM~\cite{ke2017lightgbm}, a gradient boosting framework that constructs an additive ensemble of decision trees. The prediction function is defined as

\begin{equation}
\hat{y} = \sum_{m=1}^{M} f_m(x)
\end{equation}

where $f_m(\cdot)$ denotes the $m$-th regression tree. Model parameters are learned by minimizing the regularized objective

\begin{equation}
\mathcal{L} = \sum_{i=1}^{N} \ell(y_i,\hat{y}_i) + \sum_{m=1}^{M} \Omega(f_m)
\end{equation}

where $\ell(\cdot)$ is the loss function and $\Omega(f)$ penalizes tree complexity.

Each NEAT genome generates ten LightGBM hyperparameters controlling model capacity, regularization, and stochastic sampling, including number of estimators, maximum depth, number of leaves, learning rate, feature fraction, bagging fraction, L1 and L2 regularization coefficients, minimum child samples, and categorical smoothing. Candidate configurations are trained using SMOTE-balanced data~\cite{chawla2002smote} and evaluated using stratified cross-validation.

\textbf{NEAT-Optimized AttentionMLP:}

The second learner is AttentionMLP, a neural architecture designed for low-dimensional tabular classification. The model begins with a feature attention mechanism inspired by~\cite{bahdanau2015neural} that reweights input descriptors:

\begin{equation}
\alpha_i = \sigma(w_i x_i + b_i)
\end{equation}

\begin{equation}
\tilde{x}_i = \alpha_i x_i
\end{equation}

where $\alpha_i \in [0,1]$ represents the attention weight assigned to feature $i$. The reweighted feature vector is then passed through fully connected layers ~\cite{hendrycks2016gaussian}

\begin{equation}
h^{(l+1)} = \text{GELU}(W^{(l)}h^{(l)} + b^{(l)})
\end{equation}

with batch normalization~\cite{ioffe2015batch} and dropout regularization~\cite{srivastava2014dropout}.

A separate NEAT population evolves six hyperparameters including hidden layer size, dropout rate, learning rate, weight decay, label smoothing, and Mixup coefficient. Training uses AdamW optimization~\cite{loshchilov2019decoupled} with cosine annealing schedules.

The NEAT-optimized models are combined using weighted probability fusion. Each classifier outputs class probabilities $P_i(c)$, and the final prediction is computed as

\begin{equation}
\hat{y} = \arg\max_{c} \sum_{i=1}^{M} w_i P_i(c)
\end{equation}

where $w_i$ denotes the optimized weight assigned to model $i$. The weights are determined using Nelder–Mead simplex optimization~\cite{nelder1965simplex} on out-of-fold predictions to maximize weighted F1-score. This fusion strategy leverages the complementary strengths of tree-based models and neural networks for tabular classification.

\subsection{MyoVision Data Acquisition and Research Analysis Tools}

\begin{figure*}[t]
  \centering
  \includegraphics[width=\linewidth]{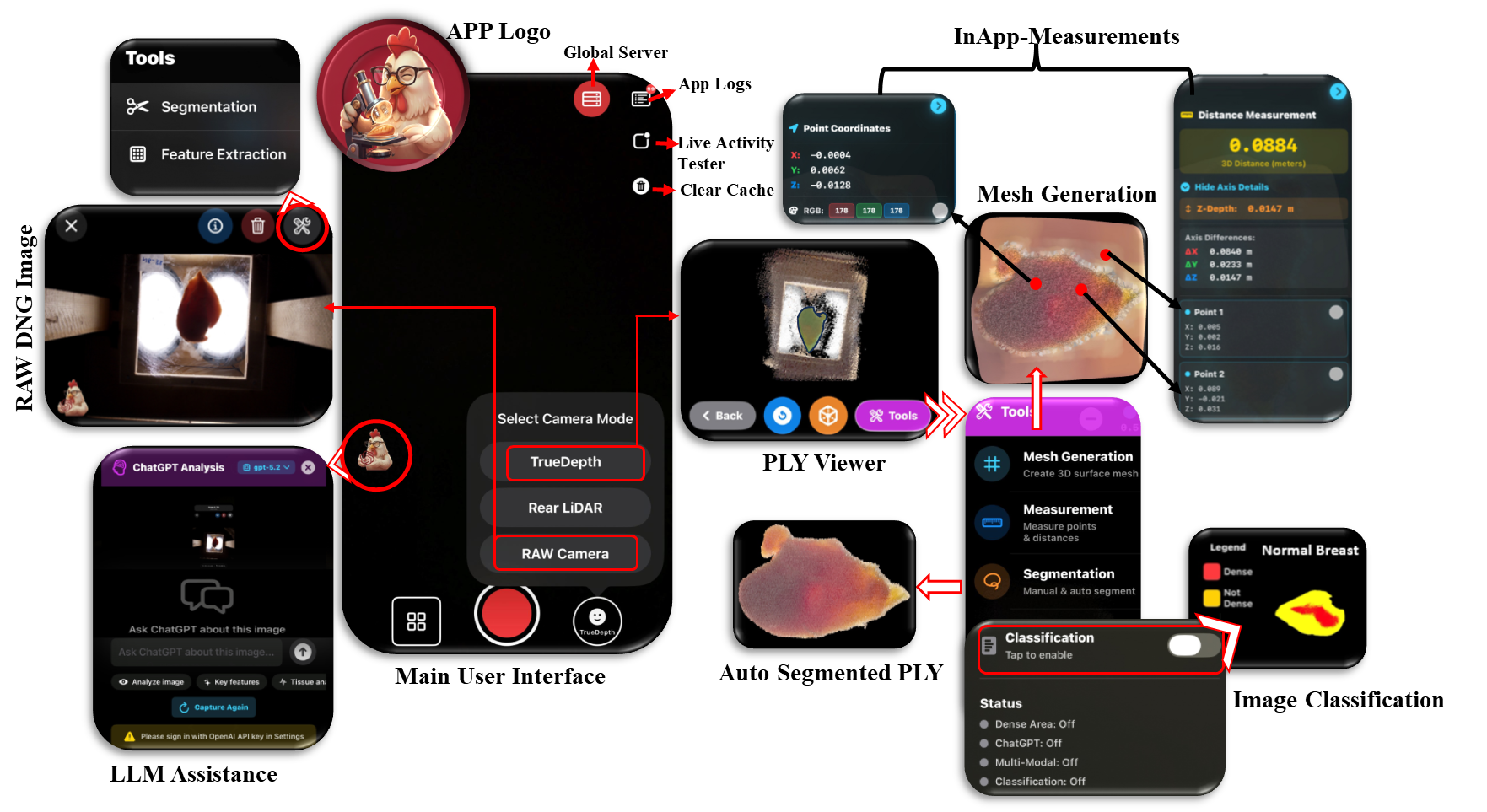}
  \caption{MyoVision Application Interface: Multi-Modal Acquisition and Analysis Pipeline.}
  \label{fig:app}
\end{figure*}

MyoVision was implemented as a native iOS application using Swift and Apple's Metal GPU framework, designed for multimodal poultry inspection data collection and real-time decision support as shown in Fig.\ref{fig:app} . The system adopts a hybrid client-server architecture: on-device components handle RAW image capture, depth sensing, and visualization, while computationally intensive tasks (mesh reconstruction, segmentation, ML inference) are offloaded to a Python-based Flask server running on workstation hardware (NVIDIA A100-PCIE-40GB GPU).

\noindent\textbf{RAW Image Acquisition.} Images are captured in 14-bit RAW (DNG) format at $4032 \times 3024$ resolution, bypassing computational photography pipelines to preserve radiometric integrity. Each capture includes a structured JSON sidecar containing over 120 metadata fields spanning photographic parameters, color science, lens calibration, IMU data, and device provenance.

\noindent\textbf{Depth Acquisition.} MyoVision supports dual-mode depth acquisition via rear-facing LiDAR (time-of-flight, $256 \times 192$, $\pm$5\,mm at 30--50\,cm) and front-facing TrueDepth (structured-light, $640 \times 480$, $\pm$1\,mm at 25--40\,cm), sharing a GPU-accelerated pipeline with $<$4\,ms per-frame latency. Adaptive hexagonal grid sampling reduces each frame to 10,000 uniformly distributed points preserving 97\% surface coverage. Depth maps are optionally reconstructed into surface meshes via Screened Poisson Surface Reconstruction~\cite{kazhdan2013screened}, enabling future multimodal structural analysis beyond the 2D classification experiments presented here.

\noindent\textbf{Research Tools.} Server-side mesh reconstruction processes raw point clouds (5--12 million points per 30--60 second recording) through a PyMeshLab pipeline performing normal estimation, implicit surface fitting, density-based trimming, Taubin smoothing, and optional quadric-edge-collapse decimation. An interactive touch-based measurement panel operates directly on reconstructed PLY meshes, supporting single-point coordinate inspection and two-point distance estimation for rapid verification of tissue thickness and depth reconstruction accuracy, with sub-50\,ms response times via GPU-based raycasting. Segmentation supports both manual polygon-based ROI definition and automated server-side processing combining RANSAC plane detection, height and color thresholding, statistical outlier removal, and SAM-based pixel-wise masking~\cite{kirillov2023segment} from a single foreground prompt.

\noindent\textbf{Feature Extraction.} From each transilluminated RAW image, 16 handcrafted spatial and frequency-domain descriptors are extracted on grayscale images. Spatial descriptors include gradient magnitude statistics (mean, std), local intensity variance, edge density, and directional gradient histograms (5 bins). Frequency-domain descriptors are computed using multi-orientation Gabor filters ($0^\circ$, $45^\circ$, $90^\circ$, $135^\circ$) to capture directional muscle fiber alignment. Morphological processing identifies dense tissue regions and computes their relative area within the fillet. The resulting descriptor vector $\mathbf{x} = [x_1, x_2, \dots, x_{16}]$ serves as input to the classification models.

\noindent\textbf{AI-Assisted Analysis.} MyoVision integrates OpenAI's ChatGPT~\cite{openai2023gpt4} via authenticated API access for context-aware user assistance, including DuckDuckGo browser access for real-time information retrieval and persistent conversation memory across interactions. Users can direct the assistant to visually analyze captured fillet images for qualitative assessment, though observations should be interpreted as supplementary guidance rather than definitive myopathy diagnoses.

\subsection{Training and Implementation Details}
\label{sec:experimental}
\begin{figure}
    \centering
    \includegraphics[width=1\linewidth]{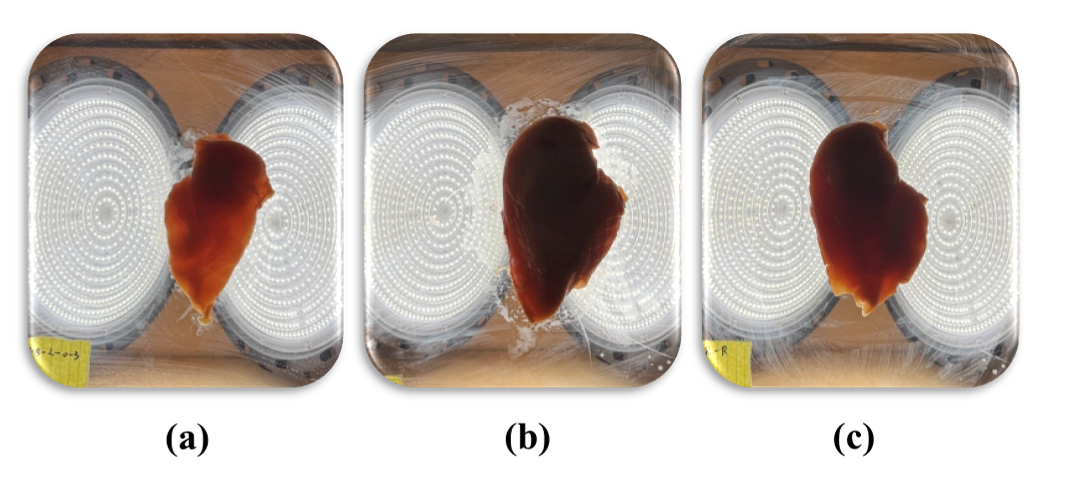}
    \caption{Transilluminated images of chicken breast fillets showing (a) normal breast, (b) woody breast, and (c) spaghetti meat. }
    \label{fig:placeholder}
\end{figure}

The complete dataset of transilluminated images shown in Fig.\ref{fig:placeholder} ($n=336$) was partitioned into a training set ($n=251$; 63 Normal, 135 WB, 52 SM), validation set ($n=34$; 9 Normal, 19 WB, 7 SM), and held-out test set ($n=51$; 13 Normal, 28 Woody Breast, 10 Spaghetti Meat). Model selection and hyperparameter optimization were conducted exclusively on the development set ($n=285$) using stratified five-fold cross-validation with out-of-fold predictions.

Within each training fold, SMOTE~\cite{chawla2002smote} oversampling balanced the three classes. Model performance was evaluated using accuracy, weighted precision, weighted recall, and macro-weighted F1-score. All experiments were repeated with fixed random seeds to ensure reproducibility. The training pipeline was implemented using PyTorch~\cite{paszke2019pytorch}, scikit-learn~\cite{pedregosa2011scikit}, Open3D~\cite{zhou2018open3d}, and PyMeshLab~\cite{kazhdan2013screened}.

%% file: sec/3_results.tex
\section{Results}
\label{sec:results}

\textbf{Experimental Protocol.}
The evaluation protocol, dataset partitions, and training procedure follow the setup described in \cref{sec:experimental}. Performance is reported using accuracy, weighted precision, weighted recall, and macro-weighted F1-score.

\subsection{Statistical Analysis of Backlighting Features}
\label{sec:stats}
To evaluate the discriminative capacity of the handcrafted descriptors extracted from transilluminated images, statistical analysis was performed on the 16-dimensional feature space. One-way ANOVA revealed that \textit{Percentage Dense Area} was the most discriminative feature across the three classes ($F = 53.90$, $p < 0.001$). Additional statistically significant descriptors included \textit{Gradient Histogram Bin~5} ($F = 31.22$, $p < 0.001$), \textit{Mean Local Variance} ($F = 31.06$, $p < 0.001$), and \textit{Standard Deviation of Gradient Magnitude} ($F = 27.53$, $p < 0.001$). In contrast, \textit{Edge Pixel Count} and \textit{Gradient Histogram Bin~2} showed no statistically significant differences across classes ($p > 0.05$).

\begin{figure}[t]
  \centering
  \includegraphics[width=\linewidth]{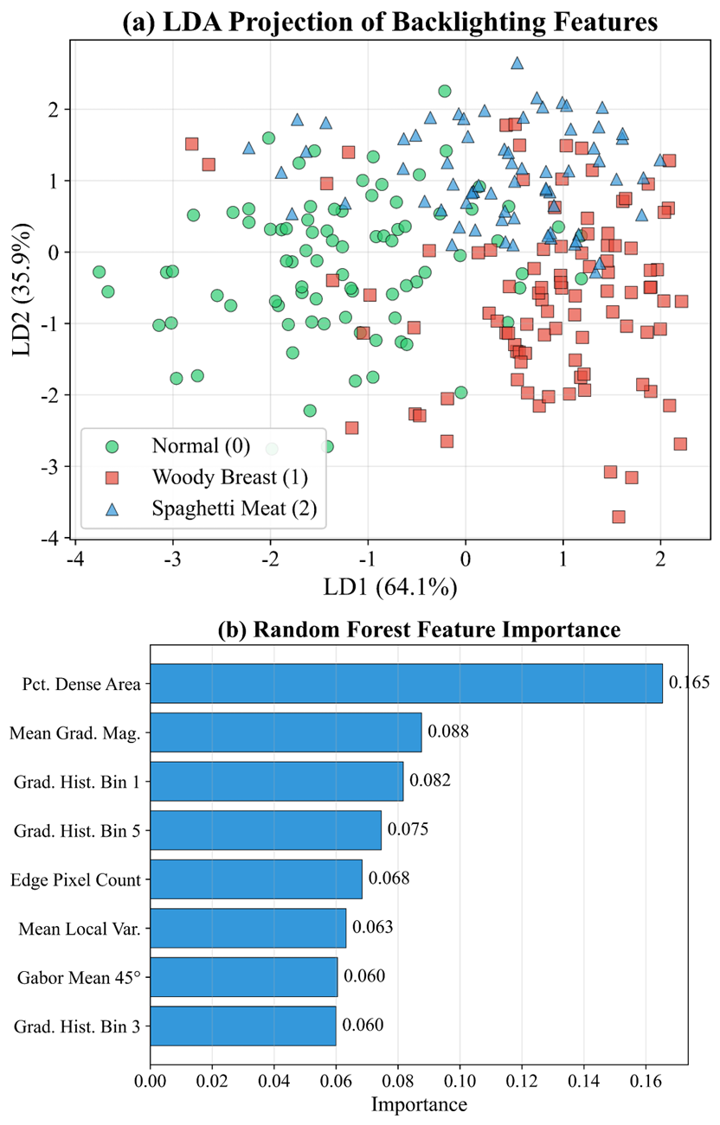}
  \caption{LDA projection of backlighting features (top) and Random Forest feature importance (bottom) for three-class myopathy classification (Normal, Woody Breast, Spaghetti Meat).}
  \label{fig:stats}
\end{figure}

To visualize separability within the feature space, Linear Discriminant Analysis (LDA) was applied (\cref{fig:stats}, a). The first discriminant component (LD1) explained 64.1\% of the variance and provided moderate separation between Normal and Woody Breast samples. However, Spaghetti Meat exhibited substantial overlap with both categories, indicating weaker linear separability.

Feature importance analysis using a Random Forest classifier (\cref{fig:stats}, b) confirmed Percentage Dense Area as the dominant predictor, followed by gradient-based descriptors capturing local texture variation and fiber orientation. A Random Forest trained directly on the raw features achieved $62.7\% \pm 17.2\%$ cross-validated accuracy, indicating that while the backlighting descriptors capture relevant structural information, the feature space alone provides limited separability for reliable three-class classification.

\subsection{Classification Performance}
\label{sec:classification_results}

Table~\ref{tab:results} summarizes the classification performance of the proposed method and four baseline models using cross-validated out-of-fold predictions and a held-out test set.

We compare against representative tabular learning models. 
\textbf{LightGBM} provides a strong tree-based baseline for structured data ~\cite{ke2017lightgbm, grinsztajn2022tree}. 
\textbf{AttentionMLP} evaluates feature-wise attention within a multilayer perceptron. 
\textbf{TabularCNN} captures local feature interactions using one-dimensional convolutions ~\cite{kiranyaz20191d}. 
\textbf{LightTransformer} adapts the Transformer self-attention mechanism to low-dimensional tabular inputs using a lightweight architecture ~\cite{vaswani2017attention}. 
These baselines span tree-based, convolutional, and attention-based learning paradigms.

\begin{table}[t]
  \caption{Combined Validation (5-Fold OOF, $n=285$) and Test ($n=51$) performance. Best results in \textbf{bold}.}
  \label{tab:results}
  \centering
  \small
  \setlength{\tabcolsep}{3pt}
  \begin{tabular}{@{}lcccc|cccc@{}}
    \toprule
    & \multicolumn{4}{c|}{\textbf{Validation (OOF)}} & \multicolumn{4}{c}{\textbf{Test}} \\
    \textbf{Model} & Acc & Prec & Rec & F1 & Acc & Prec & Rec & F1 \\
    \midrule
    \rowcolor{green!10}
    NEATBoost-Attn.& \textbf{0.78}& \textbf{0.79}& \textbf{0.78}& \textbf{0.79}& \textbf{0.82}& \textbf{0.85}& \textbf{0.82}& \textbf{0.83}\\
    LightGBM       & 0.74& 0.75& 0.74& 0.75& 0.79& 0.80& 0.78& 0.79\\
    AttentionMLP   & 0.75& 0.77& 0.75& 0.76& 0.77& 0.82& 0.77& 0.78\\
    TabularCNN     & 0.70& 0.77& 0.70& 0.72& 0.75& 0.85& 0.75& 0.77\\
    LightTransformer& 0.74& 0.79& 0.74& 0.77& 0.73& 0.77& 0.73& 0.74\\
    \bottomrule
  \end{tabular}
\end{table}

Overall, the proposed NEATBoost-Attention ensemble achieves the best performance across all evaluation metrics, improving weighted F1-score by approximately 4–7\% over individual baseline models on the validation set and maintaining consistent gains on the held-out test set.

On the test set, the ensemble achieved an accuracy of \textbf{82.4\%} with a weighted F1-score of \textbf{0.83}. The model correctly classified 11 of 13 Normal samples, 24 of 28 Woody Breast samples, and 7 of 10 Spaghetti Meat samples.

The row-normalized confusion matrices in Fig.\ref{fig:cm} provide a detailed view of the ensemble's classification behavior. Woody Breast samples were identified most reliably, achieving the highest recall among the three classes, which reflects the stronger structural signature captured by transillumination features associated with increased tissue density and rigidity. In contrast, Spaghetti Meat remained the most challenging category, with a recall of approximately 70\% on the test set. Misclassifications primarily occurred between Spaghetti Meat and the other two classes, indicating that its intermediate structural characteristics produce feature patterns that partially overlap with both Normal and Woody Breast samples.
\begin{figure}[t]
  \centering
  \includegraphics[width=\linewidth]{figures/fig5_confusion.jpg}
  \caption{Row-normalized confusion matrices of the NEATBoost-Attention ensemble.
  Left: held-out test set ($n=51$). Cells show percentage and sample counts.
  Right: cross-validated validation predictions (OOF, $n=285$).}
  \label{fig:cm}
\end{figure}

These results are consistent with the earlier feature-space analysis (Fig.\ref{fig:stats}). The LDA projection indicated that Spaghetti Meat samples exhibit substantial overlap with both Normal and Woody Breast clusters, suggesting weaker separability in the handcrafted descriptor space. In contrast, Woody Breast samples form a more distinct structural pattern due to increased tissue density and fiber rigidity. The improved performance of the proposed ensemble therefore likely arises from its ability to capture nonlinear interactions among gradient-based and texture-based descriptors that are not easily separable using linear projections or shallow models.

Inspection of the confusion matrices reveals that most misclassifications occur between Spaghetti Meat and the other two categories. This behavior likely arises from the intermediate structural characteristics of Spaghetti Meat tissue, which can exhibit partial fiber separation without the dense rigid structure typical of Woody Breast. As a result, the corresponding transillumination texture patterns may resemble either Normal or Woody Breast samples depending on the severity of the condition. These observations suggest that incorporating complementary sensing modalities, such as depth-derived structural descriptors or hyperspectral information, may further improve discrimination of subtle myopathy variants.

%% file: sec/4_discussion.tex
\section{Discussion}
\label{sec:discussion}

The performance of the proposed \textbf{NEATBoost-Attention ensemble model} should be compared with previous studies that used image-based methods to detect poultry myopathies. Binary woody breast classification using industrial imaging systems typically reports higher accuracies. For example, Geronimo \etal~\cite{geronimo2019computer} achieved 91.8\% accuracy using a computer vision system combined with SVM classification, while Yoon \etal~\cite{yoon2022development} reported 97.4\% accuracy using a custom side-view imaging setup. Similarly, Caldas-Cueva \etal~\cite{caldascueva2021image} achieved 91\% accuracy across 900 carcasses using digital camera-based morphometric analysis. However, these studies benefit from both binary classification task and the use of dedicated industrial imaging hardware. A more meaningful comparison involves multi-class myopathy classification, where performance typically decreases due to the increased difficulty of separating multiple structural abnormalities. Kato \etal~\cite{kato2019white} reported 86.4\% accuracy for white striping severity grading using SVM, and Olaniyi \etal~\cite{olaniyi2023structured} achieved 83.4\% accuracy using structured illumination reflectance imaging combined with DenseNet121. Mu\~noz-Lapeira \etal~\cite{munozlapeira2025hyperspectral} reported 76.1\% accuracy using VIS--NIR hyperspectral imaging for multi-myopathy classification including spaghetti meat detection. The 82.4\% accuracy achieved in the present study therefore falls competitively within the 75--86\% range reported for three-class myopathy classification while exceeding the most directly comparable hyperspectral benchmark by more than 6

Beyond classification accuracy, this work contributes a portable imaging framework for poultry quality assessment. Prior studies have relied on specialized hardware such as industrial cameras, near-infrared spectroscopy systems, or hyperspectral sensors, imposing substantial cost and deployment barriers~\cite{wold2019near, yoon2022development, munozlapeira2025hyperspectral}. The proposed approach demonstrates that consumer smartphone transillumination imaging can capture useful internal structural information for myopathy detection without specialized optical hardware. The integration of NEAT-based neuroevolution further provides a flexible hyperparameter optimization strategy for ensemble models, enabling diverse model configuration discovery without manual search a property particularly valuable for small, heterogeneous agricultural datasets.

Several limitations should be acknowledged. The relatively lower spaghetti meat classification performance reflects the inherent difficulty of distinguishing this condition from normal and woody breast phenotypes using 2D transillumination features alone, as spaghetti meat samples exhibit substantial feature-space overlap with other classes. Incorporating complementary modalities particularly 3D geometric descriptors from MyoVision's LiDAR-based point cloud pipeline could enable multimodal learning capturing both internal tissue structure and external surface morphology. The relatively small dataset, particularly for spaghetti meat, also limits generalization; multi-facility data collection would improve decision boundary robustness. From a deployment perspective, the current setup requires a controlled backlighting source, and future work could investigate adaptive exposure control or conveyor-mounted illumination for inline processing environments. Beyond myopathy detection, the modular MyoVision platform could extend to white striping grading, bruise detection, skin color evaluation, and pale-soft-exudative identification, supporting multiple poultry quality inspection tasks on a single mobile platform.

%% file: sec/5_conclusion.tex
\section{Conclusion}
\label{sec:conclusion}
This work demonstrated that smartphone-based transillumination imaging combined with a NEAT-optimized ensemble classifier can effectively detect poultry breast myopathies. The proposed NEATBoost-Attention framework achieved 82.4\% test accuracy (F1 = 0.83) for three-class classification of Normal, Woody Breast, and Spaghetti Meat fillets, performing competitively with prior approaches that rely on specialized imaging hardware. The results show that transillumination-derived texture descriptors capture meaningful internal structural differences between myopathy classes, while the NEAT-optimized ensemble improves classification robustness on small tabular datasets. Beyond classification performance, the MyoVision platform demonstrates the feasibility of using consumer smartphone sensors as a practical data acquisition and analysis tool for poultry quality assessment. Future work will focus on integrating multimodal features, including 3D geometric descriptors and larger multi-facility datasets, to further improve detection of subtle structural abnormalities such as spaghetti meat.
\section*{Acknowledgement}
{\sloppy
The authors acknowledge support from the University of Arkansas Experiment Station and College of Engineering, USDA NIFA (Grant Nos.\ 2023-70442-39232 and 2024-67022-42882), and NSF (Grant No.\ 2542318). Computing resources by the Arkansas HPCC, funded by NSF and the Arkansas Economic Development Commission.
}